\documentclass[letterpaper, 10 pt, conference]{ieeeconf}  

\IEEEoverridecommandlockouts                              

\usepackage{graphicx} 
\usepackage{subfigure}
\usepackage{hyperref}

\usepackage{amsmath} 
\usepackage{amsfonts,amssymb}  

\title{\LARGE \bf
MetaGrasp: Data Efficient Grasping by Affordance Interpreter Network
}

\author{Junhao Cai$^{1}$, Hui Cheng$^{1*}$, Zhanpeng Zhang$^2$ and Jingcheng Su$^1$
\thanks{This work is supported by Major Program of Science and Technology Planning Project of Guangdong Province (2017B010116003), NSFC-Shenzhen Robotics Projects (U1613211), and Guangdong Natural Science Foundation (1614050001452).}
\thanks{$^1$Junhao Cai, Hui Cheng and Jingcheng Su are with School of Data and Computer Science, Sun Yat-sen University, Guangzhou, China}
\thanks{$^2$Zhanpeng Zhang is with SenseTime Group Limited}
\thanks{*Corresponding author: chengh9@mail.sysu.edu.cn}
\thanks{More details will be released at \href{https://sysu-robotics-lab.github.io/MetaGrasp}{https://sysu-robotics-lab.github.io/MetaGrasp}}
}

\begin{document}

\maketitle
\thispagestyle{empty}
\pagestyle{empty}

\begin{abstract}

Data-driven approach for grasping shows significant advance recently. But these approaches usually require much training data. To increase the efficiency of grasping data collection, this paper presents a novel grasp training system including the whole pipeline from data collection to model inference. The system can collect effective grasp sample with a corrective strategy assisted by antipodal grasp rule, and we design an affordance interpreter network to predict pixelwise grasp affordance map. We define graspability, ungraspability and background as grasp affordances. The key advantage of our system is that the pixel-level affordance interpreter network trained with only a small number of grasp samples under antipodal rule can achieve significant performance on totally unseen objects and backgrounds. The training sample is only collected in simulation. Extensive qualitative and quantitative experiments demonstrate the accuracy and robustness of our proposed approach. In the real-world grasp experiments, we achieve a grasp success rate of 93\% on a set of household items and 91\% on a set of adversarial items with only about 6,300 simulated samples. We also achieve 87\% accuracy in clutter scenario. Although the model is trained using only RGB image, when changing the background textures, it also performs well and can achieve even 94\% accuracy on the set of adversarial objects, which outperforms current state-of-the-art methods.

\end{abstract}

\section{INTRODUCTION}

Recently vision-based data-driven grasp synthesis has achieved significant performance on robotic grasping task. Many works have shown that deep learning models trained using data labelled by human or collected by trial and error in physical environment perform well for seen and novel objects~\cite{lenz2015deep,johns2016deep,pinto2016supersizing,levine2018learning,mahler2017dex,viereck2017learning,zeng2017robotic,morrison2018closing,kalashnikov2018qt}.

However, manually annotating data~\cite{zeng2017robotic,morrison2018closing} and doing trial and error in real environment~\cite{pinto2016supersizing,levine2018learning,kalashnikov2018qt} are time-consuming. An alternative way to alleviate this issue is to generate data from virtual environment, while much of simulated data cannot be directly used because of the domain shift problem~\cite{sugiyama2017dataset}. Therefore, some approaches used simulated RGB data and domain adaptation to reduce the number of real-world samples and achieve comparable performance, while the number of real-world samples required is still large~\cite{bousmalis2017using,fang2017multi}.

\begin{figure}[t]
    \begin{center}
        \includegraphics[scale=0.26]{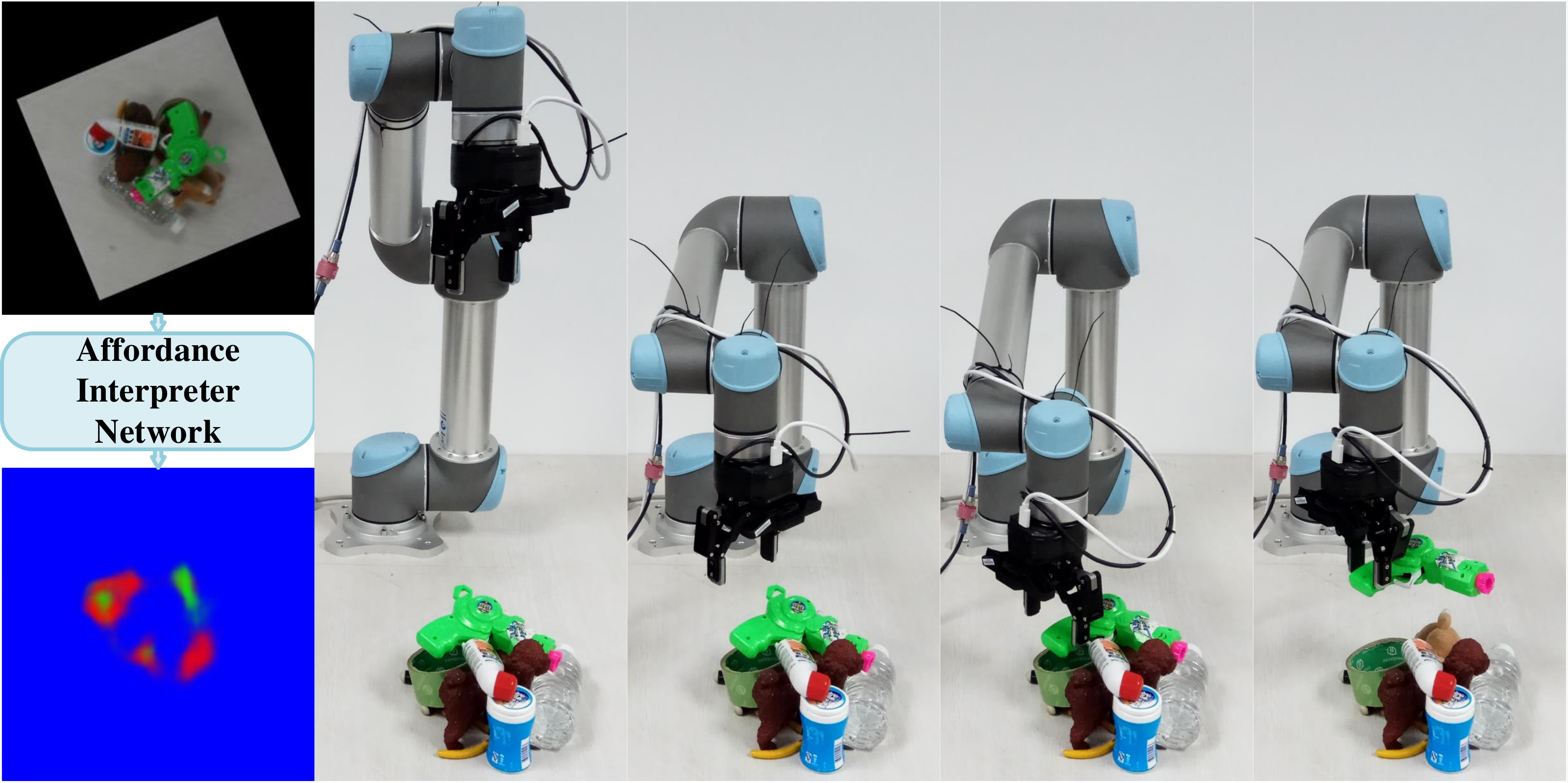}
    \end{center}
    \vspace{-10pt}
    \caption{{\bf{The proposed grasp system pipeline.}} Given a global workspace RGB image, the model predicts horizontal grasp affordance maps with respect to different rotations in the camera's reference frame. Unlike~\cite{zeng2017robotic}, we directly obtain pixelwise affordance maps with only RGB image instead of upsampling the model output. The greenest region represents the most stable horizontal grasp point, red region stands for negative grasp location and blue is background. The right side shows a grasp episode.}
\label{fig:grasp system pipeline}
\vspace{-20pt}
\end{figure}

On the other hand, many approaches have shown that models trained by only synthesized depth data work well in physical world~\cite{mahler2017dex,viereck2017learning,tobin2017domainr}. Specifically, the promising grasping performance achieved by~\cite{mahler2017dex,mahler2017learning} partially give the credit to using antipodal grasp sampling method in depth image~\cite{goldberg1999part}, this grasp rule is able to generate robust grasp samples. However, in the real-world setting, the performance degenerates when trying to grasp tiny, meshed and translucent objects, because the depth information is missing for these items due to their special physical properties. Moreover, the camera configurations such as camera height corresponding to the workspace in simulator and real world should be consistent. In addition, the number of the training data in~\cite{mahler2017dex} is extremely huge, and the collected samples in simulator only contain local depth image of which the information is limited, hence they cannot utilize the context of global visual information to facilitate the grasp planning.

 Therefore, we turn to explore another kind of grasp training system that captures grasp patterns only in RGB image. The system can effectively utilize grasp samples collected from simulator. In our proposed system, the generated grasp samples are based on antipodal grasp heuristic instead of random grasping as in~\cite{pinto2016supersizing,levine2018learning,kalashnikov2018qt}. Also, we use global RGB visual-based perception, not the local depth data as in~\cite{mahler2017dex}. A well-interpreted model is proposed to evaluate accurate grasp pattern trained using only a small number of collected samples without any real-world data.

For the data collection system, we propose a novel way to generate simulated data guided by antipodal grasp rule. Unlike other self-supervised methods which collect data only by random grasping~\cite{pinto2016supersizing,levine2018learning,kalashnikov2018qt}, we use a corrective grasp strategy to generate antipodal grasp samples without cumbersomely sampling method~\cite{mahler2017dex}. The data collected by our system emphasizes the antipodal grasp pattern and has less perturbed information than data generated by random grasping trials, which is beneficial to training the model. In our system, we only collect global RGB image containing the whole workspace, object positions in the image and corresponding grasp label. We show that we can directly apply the model trained with only about 6,300 simulated data (approximately 1000 times smaller than~\cite{mahler2017dex}) to the real-world grasping task and achieve equivalent performance compared with current state-of-the-art techniques.

For the grasp inference model, we employ an end-to-end model like~\cite{zeng2017robotic} to learn grasp meta knowledge. Differently we extend the architecture to directly predict pixelwise grasp affordance map without the upsampling operation on the output. In this work, we define graspability, ungraspability and background as grasp affordances, i.e., each pixel of affordance map belongs to one of grasp affordances. In addition, we train the network with a novel loss function to better do with label sparsity problem. Extensive experiments show that our model can predict more accurate affordance map and achieve higher performance compared to~\cite{zeng2017robotic}.

In order to test the performance of our system, we perform grasp tries and evaluate the grasp success rate in several different scenarios, where we achieve significant grasp results. The grasp items are totally unseen for the model. To demonstrate the robustness of our approach, we also make grasp attempts on the set of adversarial items~\cite{mahler2017dex} under different background textures. The model is still able to predict the correct grasp configuration, which achieves even 94\% success rate in a background texture that totally different to that of the simulated background.

In summary, our contributions mainly are:
\begin{itemize}
\item Designing a data collecting system guided by the antipodal grasp rule in virtual environment. The collected sample is composed of the RGB image containing the whole workspace where there is only a single object placed, the object positions corresponding to the image and the grasp labels. We show that we can achieve significant performance using only a small number of samples without any real-world data.
\item We construct an end-to-end affordance interpreter network to achieve pixel-level grasp configuration and demonstrate that the network is able to learn antipodal grasp pattern from extremely sparse labeled dataset containing only about 6,300 samples without extensive domain randomization (i.e., we fix background, gripper parameters, camera location and lighting condition, and use only 35 objects from 3DNet~\cite{wohlkinger20123dnet}).
\item We evaluate our model performance on extensive experiments that consist of grasping unseen household and adversarial objects~\cite{mahler2017dex} in different scenarios, which shows better performance compared to current state-of-the-art methods.
\end{itemize}

\section{RELATED WORK}

\noindent{\bf{Grasp Planning.}} Robotic grasping is one of the most widely studied topics in the area of object manipulation. In grasping task, we aim to achieve a desired object constraint in front of external disturbance~\cite{suarez2006grasp}. The existing grasping techniques can be divided into analytic and empirical methods. Analytic methods try to evaluate grasp configuration according to some metric~\cite{ferrari1992planning,nguyen1988constructing,weisz2012pose,pokorny2013classical} on the assumption of simplified contact model, Coulomb friction and rigid-object modeling~\cite{siciliano2016springer}. While these methods often required known object model and location~\cite{prattichizzo2008grasping}. Contrary to analytic approaches, empirical methods try to obtain object representations from data, which can be used to evaluate grasp configuration with heuristics~\cite{bohg2014data}.

\noindent{\bf{Deep Learning for Grasping.}} Because of the success of deep learning in visual perception~\cite{he2016deep,long2015fully,ren2015faster}, more and more grasping system make use of a vision-based deep learning framework to predict grasp configuration~\cite{lenz2015deep,johns2016deep,pinto2016supersizing,levine2018learning,mahler2017dex,viereck2017learning,zeng2017robotic,morrison2018closing,kalashnikov2018qt}. Levine \emph{et al.} spent two months on collecting 800k grasp samples using 6-14 robots to train an evaluation network~\cite{levine2018learning}. Zeng \emph{et al.} proposed an end-to-end network to predict grasp affordance map using manually annotated data~\cite{zeng2017robotic}. Morrison \emph{et al.} utilized Cornell grasping dataset to train a light-weight convolutional neural network to achieve pixelwise grasp pose prediction~\cite{morrison2018closing}. These methods achieve significant performance, while data collected from hand-designed procedure or pure trial-and-error manner in real physical environment is time-consuming and less effective, which reduce the practicability.

\noindent{\bf{Virtual to Real-world Transfer in Grasping.}} Because we can easily have access to a large amount of annotated data from virtual environment, many works turned into simulator to generate training samples, especially in object manipulation task~\cite{tobin2017domain,pinto2017asymmetric,peng2017sim,matas2018sim}. In grasping scenario, Viereck \emph{et al.} trained a distance function between current pose and nearest optimal pose using only simulated data~\cite{viereck2017learning}. Mahler \emph{et al.} obtain synthesized data based on antipodal grasping sampling method and trained a grasp quality network to evaluate robust grasp configuration~\cite{mahler2017dex}. All of them use only simulated data and work well in real physical system, a key factor is the visual related data collected from simulator is depth data instead of RGB image. Viereck \emph{et al.} believed that depth image contains less information than that of RGB image~\cite{viereck2017learning}. Hence it is easier to bridge the gap in transferring models trained in a simulator to the physical world. However, in the physical world setting, it is difficult for depth camera to measure thin, translucent and dark color objects due to their special physical properties. Under this circumstance, the grasp performance is no longer guaranteed.

\noindent{\bf{Approach of Data Collection for Grasping.}} Large amount of available annotated data is one of the key factors for the success of deep learning for grasping. The works in ~\cite{pinto2017asymmetric,levine2018learning,kalashnikov2018qt,fang2017multi,bousmalis2017using} proposed trial-and-error ways in real world to collect data. Although most works
achieve good results, the number of required samples is relatively large (from 50k to 800k). One of the most important reason is that the data collection systems execute grasp trials without any heuristic. Dex-net 2.0 generated grasp samples with antipodal grasp sampling approach~\cite{goldberg1999part} to obtain robust grasp~\cite{mahler2017dex}. However, they only collect local depth information, which needs 6.7 million samples to train the model. Zeng \emph{et al.} collected global visual samples and annotated them manually, it is less efficient and error-prone~\cite{zeng2017robotic}. We believe that better grasp pattern can be obtained from global visual perception which can dramatically reduce the required number of training samples. Therefore, a better data collection system should be able to efficiently do grasp trials based on specific rule and obtain global visual information.

\section{APPROACH}

\subsection{Problem Formulation}

Similar to the related literature~\cite{pinto2016supersizing,mahler2017dex,morrison2018closing}, given an RGB image, we consider the grasping problem as executing parallel-jaw grasp in planar workspace.

Let $g = (p,\omega, \eta)$ defines a perpendicular parallel-jaw grasp, where $p = (x,y,z)$ denotes the location of the grasp point in Cartesian, $\omega \in [0, 2\pi)$ represents the grasp angle with respect to the end effector and $\eta \in \mathbb{R}^3$ is a one hot vector which indicates the grasp affordances. When projected to image space, the grasp in image $I$ can be represented as $\hat{g} = (\hat{p}, \hat{\omega}, \eta)$, where $\hat{p} = (u,v)$ depicts the grasp location in image, $\hat{\omega} \in \{\frac{\pi}{8} \times i \mid i = 0,..., 15\}$ represents discretized grasp angle. The discretization operation can reduce the complexity of learning procedure (note that in the inference process, we can replace the rotation of gripper equivalently by rotating the input image. Therefore, we can consider only the horizontal grasp for the image in the training process).
Hence, we can define grasp affordance for each image pixel, and the grasp affordance maps can be written as: $A = [C_0, C_1,..., C_{15}] \in \mathbb{R}^{16 \times H \times W \times 3}$, where $C_i \in \mathbb{R}^{H \times W \times 3}$ denotes the grasp affordances for the whole image given the condition of angle $i$. The 3 channels here mean the positive, negative grasp and background. According to the affordance maps, we can extract the first channel $C_{i,0} \in \mathbb{R}^{H \times W}$ (0 indices the positive affordance map in $C_i$) from each $C_i$ given angle $i$ and compose them together as : $G = [C_{0,0}, C_{1,0},..., C_{15,0}] \in \mathbb{R}^{16 \times H \times W}$. Therefore, the most robust grasp configuration can be evaluated by
\vspace{-0.5\baselineskip}
\begin{equation}
i^*,h^*,w^* = \mathop{\arg\max}_{i,h,w} G(i,h,w)
\vspace{-0.5\baselineskip}
\end{equation}
where $G(i,h,w)$ denotes value of the positive affordance given the position of pixel and rotated angle. The pair $(h^*,w^*)$ is the location where end-effector should reach in image space, and $i^*$ represents that the gripper rotates $i^* \times \frac{\pi}{8}$ degrees before executing grasp.

In training process, we define a parameterized function $f_\theta$ to achieve pixel-level mapping, which is denoted as:
\vspace{-0.5\baselineskip}
\begin{equation}
C_{o_i} = f_\theta(I_{r_i})
\vspace{-0.5\baselineskip}
\end{equation}
where $I_{r_i}$ is the image obtained by rotating $i \times \frac{\pi}{8}$ degrees with respect to $I$, $C_{o_i}$ is the output affordance map corresponding to $I_{r_i}$, and $f_\theta$ is implemented using a neural network. Applying the softmax loss function $L$, our training objective can be formulated as:
\vspace{-0.5\baselineskip}
\begin{equation}
\theta^* = \mathop{\arg\min}_{\theta} L(C_{o_i}, Y)
\label{eqn: objective function}
\vspace{-0.5\baselineskip}
\end{equation}
where $Y \in \mathbb{R}^{H \times W \times 3}$ represents the label map. Details are discussed in Sec.~\ref{training process}

\begin{figure}[t]	
    \centering
    \subfigure[]{
		\includegraphics[scale=0.47]{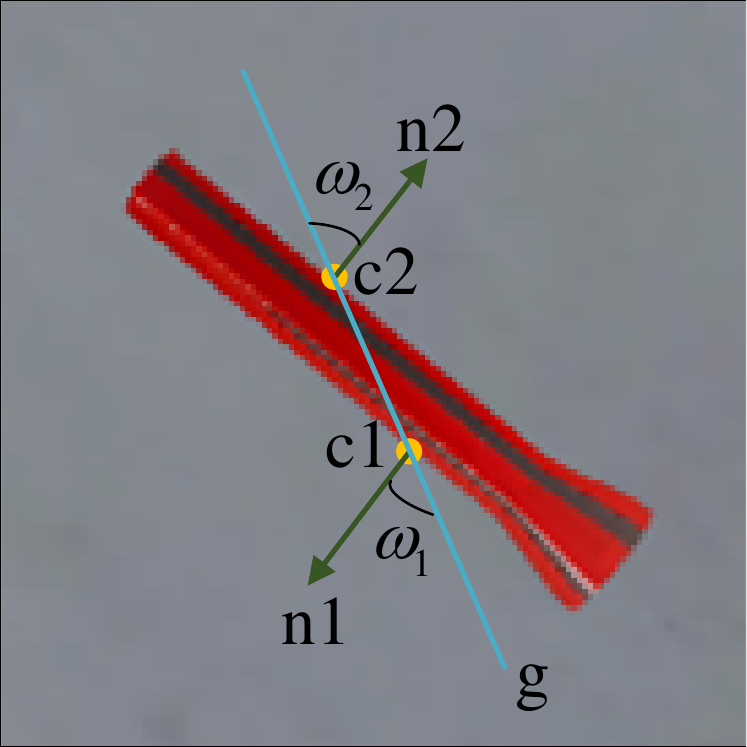}
    }
    \hspace{10pt}
    \subfigure[]{
		\includegraphics[scale=0.47]{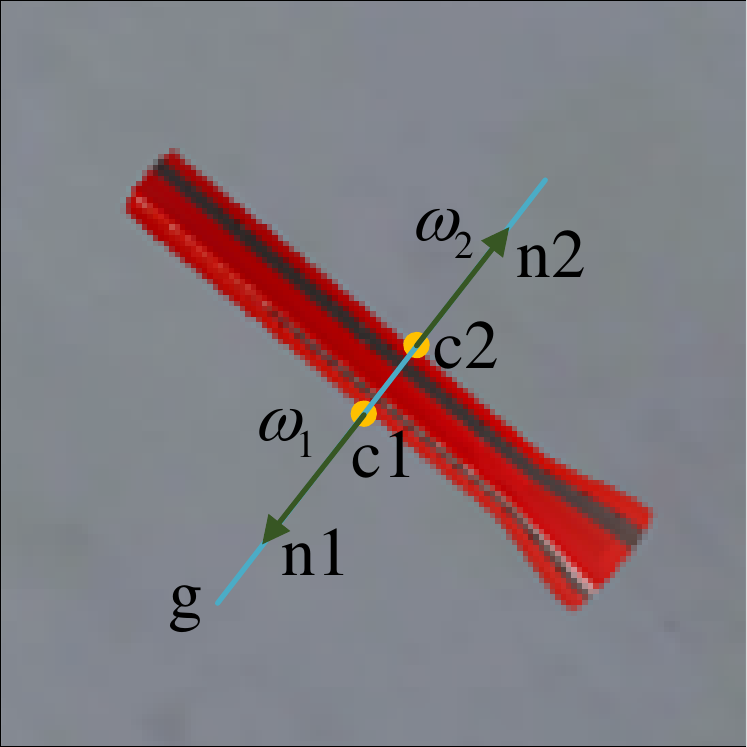}
    }
    \vspace{-10pt}
	\caption{{\bf{Antipodal grasp in simulator.}} $c1$ and $c2$ (yellow points) denote contact points between parallel-jaw gripper and rigid object. $n1$ and $n2$ (green directed lines) is the normal vector with respect to the contact points. $g$ (blue line) represents the grasp angle in image space. (a) The grasp direction is not parallel to the normal vectors, though it may be successfully grasped by some gripper with wide width, it is not a robust grasp. (b) The grasp direction is exactly parallel to the normal vectors, we consider it as a robust antipodal grasp.}
\label{fig:antipodal grasp}
\vspace{-20pt}
\end{figure}

\vspace{-4pt}
\subsection{Data Collection from Simulator}
{\bf{Definition of antipodal grasp in image space.}}
Similar to ~\cite{mahler2017dex}, we formulate antipodal grasp in image space as follows. We consider the scenario that the robot executes grasp trials in workspace where there is only an object placed. We define $c1$ and $c2$ as contact points between parallel-jaw gripper and object, $n1$ and $n2$ as normal vector for each contact point, and $g$ as grasp direction in image space. $c1$, $c2$, $n1$, $n2$ and $g$ $\in \mathbb{R}^2$ are all in the 2-D image space illustrated in Fig.~\ref{fig:antipodal grasp}. As we can see,
\vspace{-0.5\baselineskip}
\begin{equation}
g = \frac{c_1 - c_2}{\Arrowvert c_1 - c_2 \Arrowvert}
\vspace{-0.5\baselineskip}
\end{equation}
where $\Arrowvert \cdot \Arrowvert$ denotes the norm. We consider a grasp as an antipodal grasp if:
\vspace{-0.5\baselineskip}
\begin{equation}
arccos(\frac{-n_1 \cdot g}{\Arrowvert n_1 \Arrowvert}) < \theta_1
\end{equation}
\vspace{-0.5\baselineskip}
\begin{equation}
arccos(\frac{n_2 \cdot g}{\Arrowvert n_2 \Arrowvert}) < \theta_1
\end{equation}
\vspace{-0.5\baselineskip}
\begin{equation}
arccos(\frac{n_1 \cdot n_2}{\Arrowvert n_1 \Arrowvert \cdot \Arrowvert n_2 \Arrowvert}) > \theta_2
\end{equation}
are all satisfied, where $\theta_1$ and $\theta_2$ are non-negative values that are prone to 0 and $\pi$ respectively. In short, a grasp can be regarded as stably antipodal grasp when grasp direction is parallel to normal vectors of contact points in image space. We will describe how to use this rule to guide our corrective strategy for grasp data collection.

\begin{figure}[b]	
\vspace{-10pt}
	\begin{center}
		\includegraphics[scale=0.21]{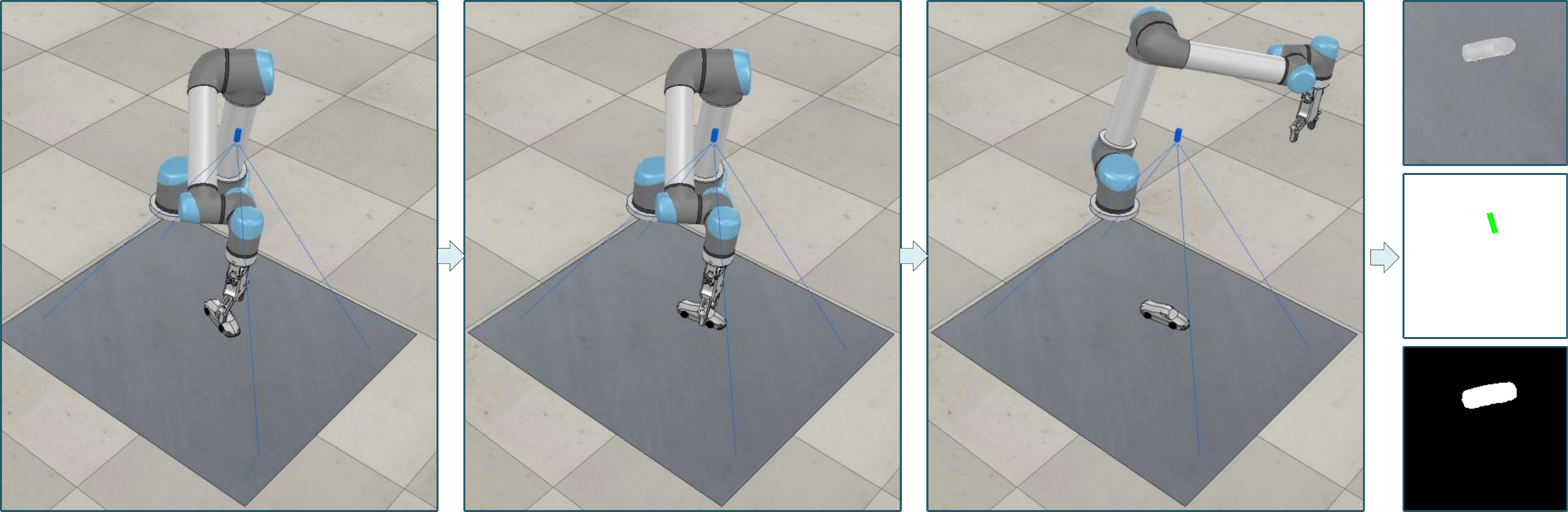}
	\end{center}
    \vspace{-10pt}
	\caption{{\bf{Illustration of data generation.}} The first three scenes on the left side shows the generation of antipodal sample by recording the global RGB image in which the object has been adjusted by the gripper. The information of saved sample contains 1) the RGB image, 2) grasp label and 3) the object mask where the white region represents locations of object.}
\label{fig:illustration_of_data_generation}
\end{figure}

\begin{figure}[t]
    \begin{center}
        \includegraphics[scale=0.23]{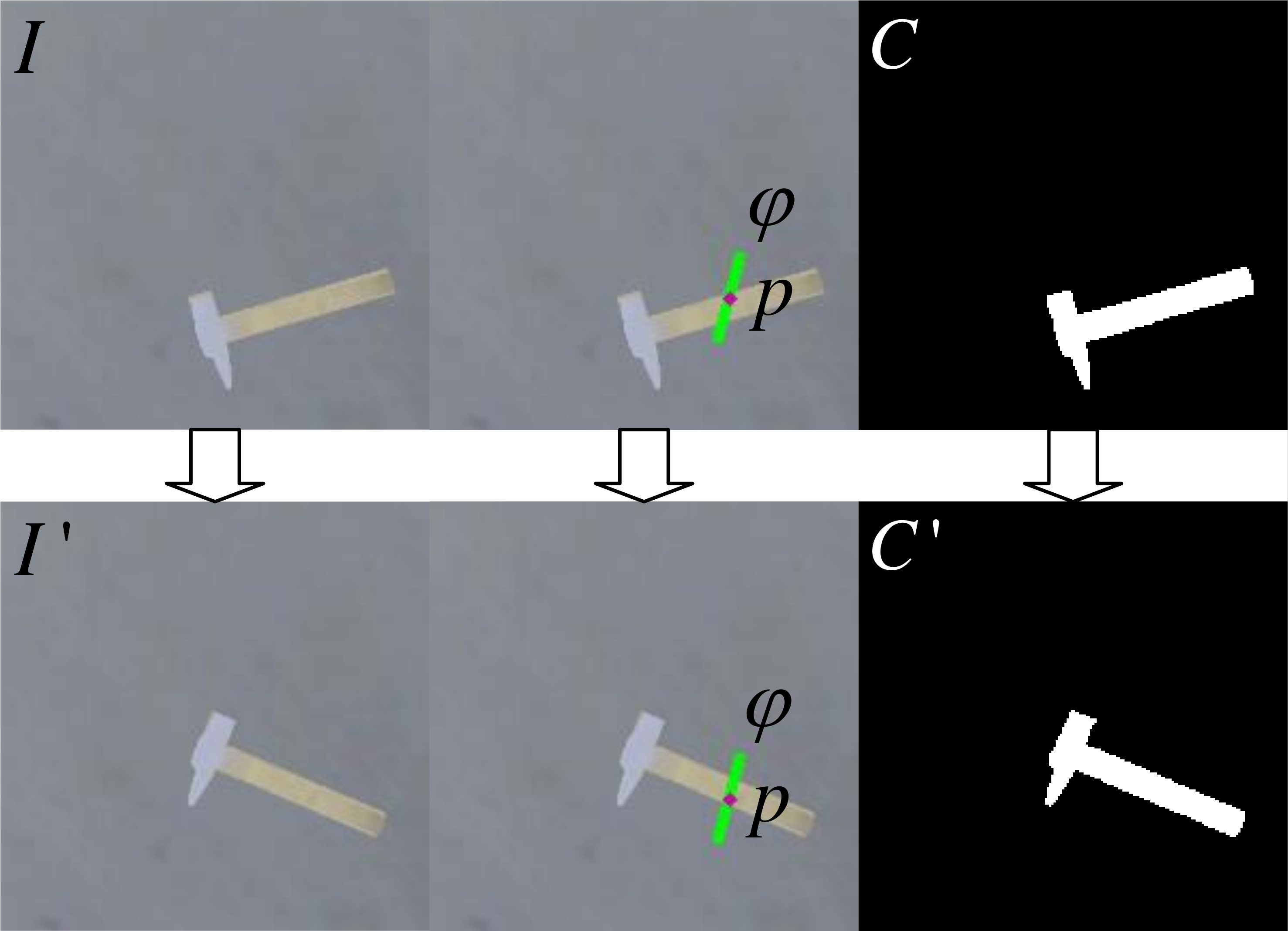}
    \end{center}
    \vspace{-10pt}
    \caption{{\bf{Sample generated by the corrective grasp strategy.}} The first column is the global RGB image captured by the camera. The green lines and red point represent the grasp angle and location respectively. And column 3 is object mask that indicate the positions of object. Row 1 displays the pattern of the sample collected before grasp trial. Row 2 shows that the grasp trial adjusts object's pose so that the contact points between object and gripper become antipodal. Therefore, we renew the pattern of sample by re-recording the global image and corresponding object mask.}
\label{fig:corrective grasp strategy}
\vspace{-20pt}
\end{figure}
{\bf{Corrective grasp strategy.}} In implementation, we directly collect antipodal grasp samples using a corrective grasp strategy. The overview of a grasp attempt is illustrated in Fig.~\ref{fig:illustration_of_data_generation}. We first randomly select a grasp angle $\varphi$ and a position $p=(u,v)$ in image space where the pixel belongs to object instead of background, and use global camera to record current RGB image $I$. We also collect all the pixel coordinates $C$ that belong to object. The angle, position, image and coordinates are saved as the pattern of a sample $S=(\varphi, p, C, I)$. Then the end-effector moves to the position $p$, rotates the angle $\varphi$ and executes grasp. If the gripper successfully grasps the object (this can be determined by the distance between two fingertips of the gripper), we open the gripper and move end-effector to the initialization position. Due to the grasp trial, the object's pose is adjusted so that the contact points between object and gripper become approximately antipodal, hence we renew the pattern of sample by using the global camera to re-record the global visual information $I^\prime$ and pixel coordinates of the object $C^\prime$, then append grasp label $l=1$ to the sample which is changed to $S=(\varphi, p, C^\prime, I^\prime; l)$. If the gripper fails to grasp the object, we simply add grasp label ($l=0$ in this case) to the sample with other information unchanged so that $S=(\varphi, p, C, I; l)$. Fig.~\ref{fig:corrective grasp strategy} shows the sample generated by corrective grasp strategy.

Note that although the method in~\cite{mahler2017dex} also employs antipodal grasp rule in the data collection process, our method is based on trial-and-error method followed by a corrective strategy. On one hand, this can avoid the limitation of the heuristic rule and explore more potential graspable points. On the other hand, the corrective strategy improves the quality of the grasp samples. In particular, the synthesized samples collected from our system effectively highlight the grasp features, hence the model can easily capture the most useful information for grasping, ignore other perturbed information from the synthesized data, and generalize to different scenes.

In this work, we construct virtual scene using the simulator V-REP~\cite{rohmer2013v}. We can effectively collect antipodal grasp sample using the above method without any sampling procedure. Specifically, we use 6DOF UR5 robot mounted with RG2 gripper, while we mount ROBOTIQ85 gripper to the wrist of UR5 in physical world. We show that different grippers between virtual and real environments have no influence on grasp performance.

{\bf{Data Structure}}
As shown in Fig.~\ref{fig:illustration_of_data_generation}, a grasp sample consists of an RGB image of the whole workspace, the grasp label that illustrates which locations are graspable or not as well as how much degrees the gripper should rotate, and object coordinates that indicate the whole object pixel locations in the RGB image.

\begin{figure}[t]
	\begin{center}
		\includegraphics[scale=0.24]{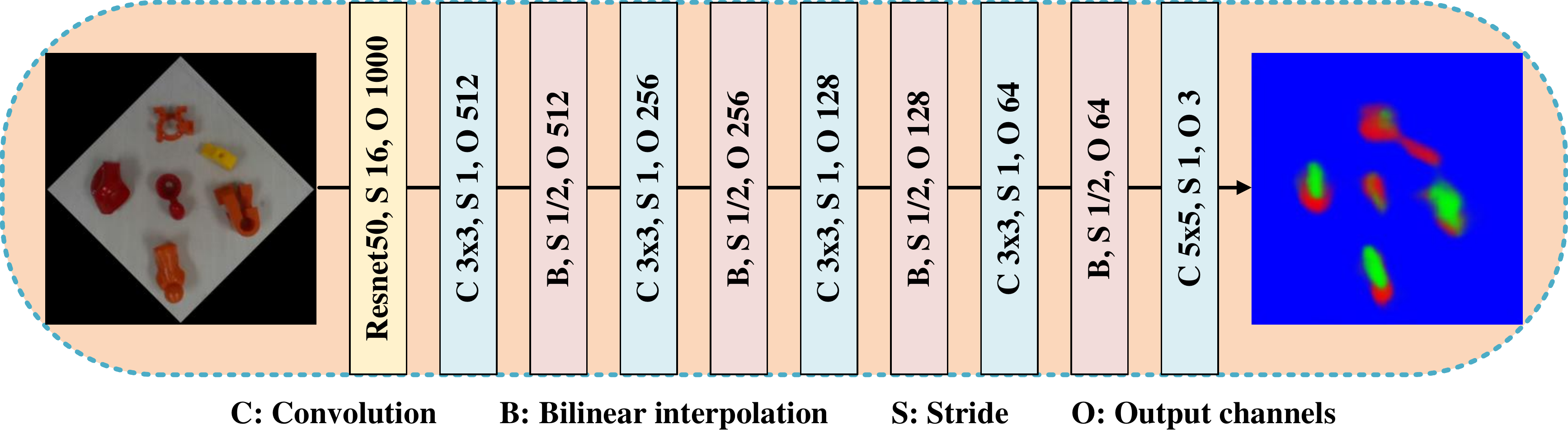}
	\end{center}
    \vspace{-10pt}
	\caption{{\bf{Affordance interpreter network.}} Given a rotated RGB image, the network outputs pixelwise affordance map in which blue region represents background, green denotes positive horizontal grasp point, red stands for negative grasp configuration.}
\label{fig:model architecture}
\vspace{-20pt}
\end{figure}

\vspace{-4pt}
\subsection{Model Architecture}
The network architecture is shown in Fig.~\ref{fig:model architecture}. Similar to~\cite{zeng2017robotic}, we design a fully convolutional residual network as affordance interpreter network to predict horizontal grasp affordance maps. Differently, in order to achieve pixelwise prediction, we append bilinear interpolation block for each 3x3 conv layer, and use 5x5 conv layer to refine affordance maps. This architecture is more accurate to evaluate graspable area than that proposed by~\cite{zeng2017robotic}.

We find that the model can avoid domain shift problem even if it is trained with only simulated RGB image. A key factor is that all the positive grasp samples are antipodal pattern based on antipodal grasp heuristic, hence the model has no need for learning other confusing grasp patterns and is able to pay more attention to the grasp meta knowledge. Under this circumstance, it is a small number of antipodal samples that is sufficient to train the model. Another key factor of this model is that it observes the global visual information so that it can predict robust results even in complex scenarios such as multiple objects and clutter.

\vspace{-4pt}
\subsection{Training Process}
\label{training process}
\begin{figure}[b]	
\vspace{-10pt}
	\begin{center}
		\includegraphics[scale=0.27]{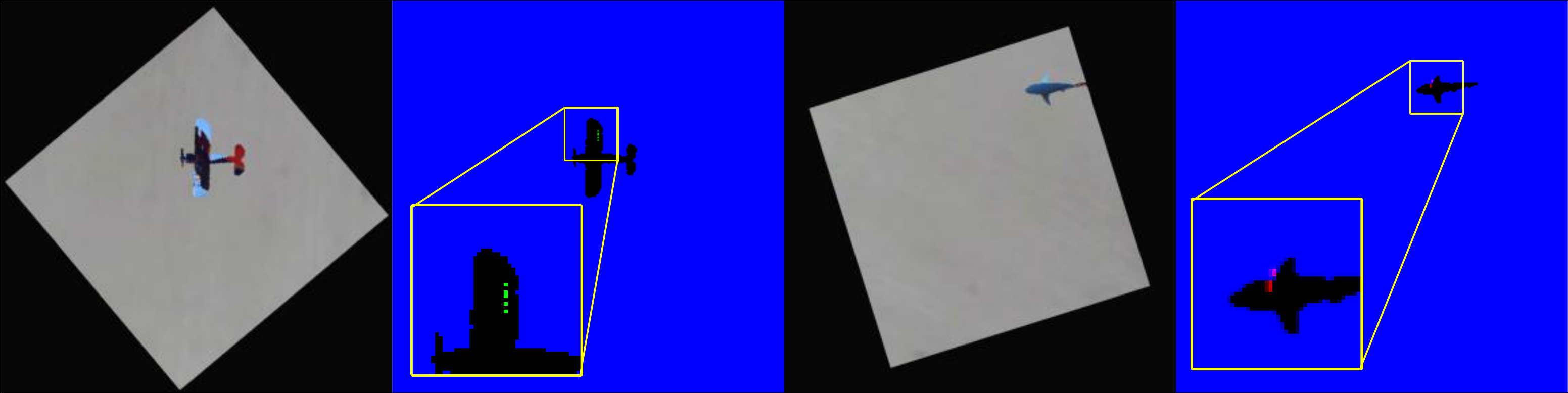}
	\end{center}
    \vspace{-10pt}
	\caption{{\bf{Label mask.}} According to collected object positions, we can generate label mask in which 0 is assigned to unlabelled object regions. We can see that there are several positive (green) or negative (red) labels within the black region.}
\label{fig:label mask}
\end{figure}
Because 1) most of the labels with respect to all the image pixels are background, 2) positive and negative grasp labels are extremely sparse with respect to the object pixels, directly training the network becomes difficult in our experiment. Therefore, we introduce the object mask $M \in \mathbb{R}^{H \times W \times 3}$ to the evaluation of loss function. As illustrated in Fig.~\ref{fig:label mask}, we set each pixel of unlabeled object region (i.e., the black region in Fig.~\ref{fig:label mask}) with $\textbf{0} \in \mathbb{R}^3$ and the others with $\textbf{1} \in \mathbb{R}^3$ to generate a training mask. Let $\phi(x) \in \mathbb{R}^{H \times W \times 3}$ denotes the feature maps of output of last convolutional layer. And the corresponding loss function of the network training is represented as:
\vspace{-0.5\baselineskip}
\begin{equation}
L = \frac{1}{3HW}\sum_{i=1}^{H}\sum_{h=1}^{W}\sum_{k=1}^{3}-Y_{ijk}M_{ijk}\log \frac{e^{\phi(x)_{ijk}}}{\sum_{l}e^{\phi(x)_{ijl}}}
\vspace{-0.5\baselineskip}
\end{equation}
where $Y \in \mathbb{R}^{H \times W \times 3}$ represents label map as in Eqn. (\ref{eqn: objective function}).

To reduce the impact of label sparsity problem, we place more weight on positive or negative label map and shrink the loss of background. Concretely, we empirically multiply positive and negative maps by scale factor 120 and background by 0.1 to adjust the influence of label with respect to the model parameters, i.e., each pixel of modified mask is formulated as:
\vspace{-0.5\baselineskip}
\begin{equation}
M^{\prime}_{ij} = [M_{ij0} \cdot 120, M_{ij1} \cdot 120, M_{ij2} \cdot 0.1]
\vspace{-0.5\baselineskip}
\end{equation}
We show that these approaches significantly reduce the difficulty of model training.

\vspace{-4pt}
\subsection{Grasp Execution}
For each grasp execution, we first capture the RGB image containing the whole workspace. Then we rotate the image into 16 orientations with different multiples of $\frac{\pi}{8}$ and feed the rotated images into the affordance interpreter network to predict horizontal grasp pixelwise affordance maps. Next, we move the end effector to the position whose corresponding image location has the highest value of graspable affordance and rotate the angle according to the orientation of the rotated image. Finally, we approach the object and execute grasp.
\section{EXPERIMENTS}

\subsection{Experimental Setting}
\textbf{Physical Component.} We use 6DOF UR5 robot mounted with ROBOTIQ85 gripper at end effector and Intel RealSense SR300 at robot's wrist. SR300 is able to capture depth information. Note that we only use depth image to obtain suitable gripper reaching height not for grasping affordance inference. The workspace contains a $40cm \times 40cm$ planar area with grey texture. We change workspace texture at the robustness experiments.

\begin{figure}[t]	
    \centering
    \subfigure[Training objects]{
        \label{fig:train objects}
		\includegraphics[scale=0.0889]{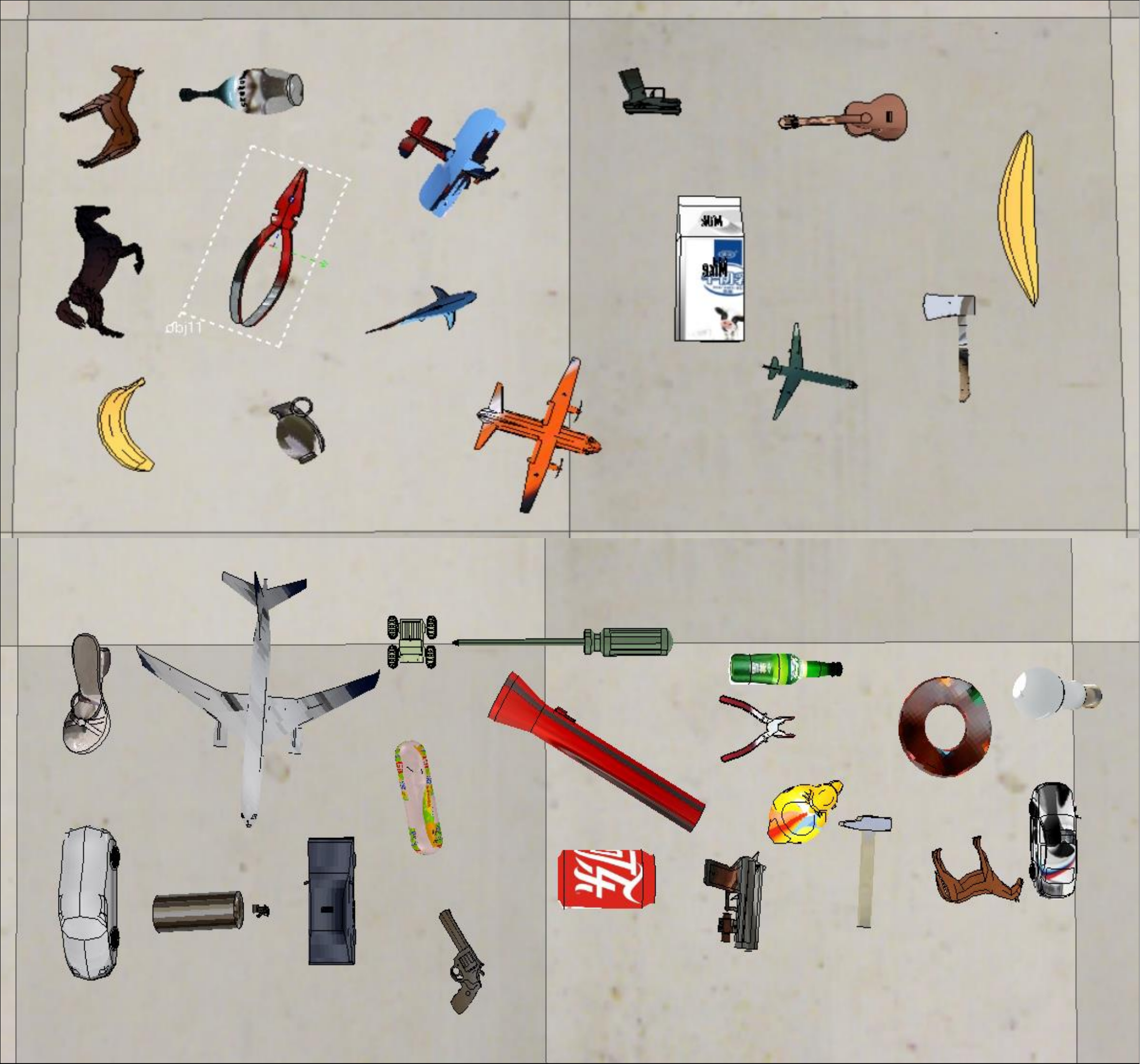}
    }
    \hspace{-11pt}
    \subfigure[Test objects]{
        \label{fig:test objects}
		\includegraphics[scale=0.0889]{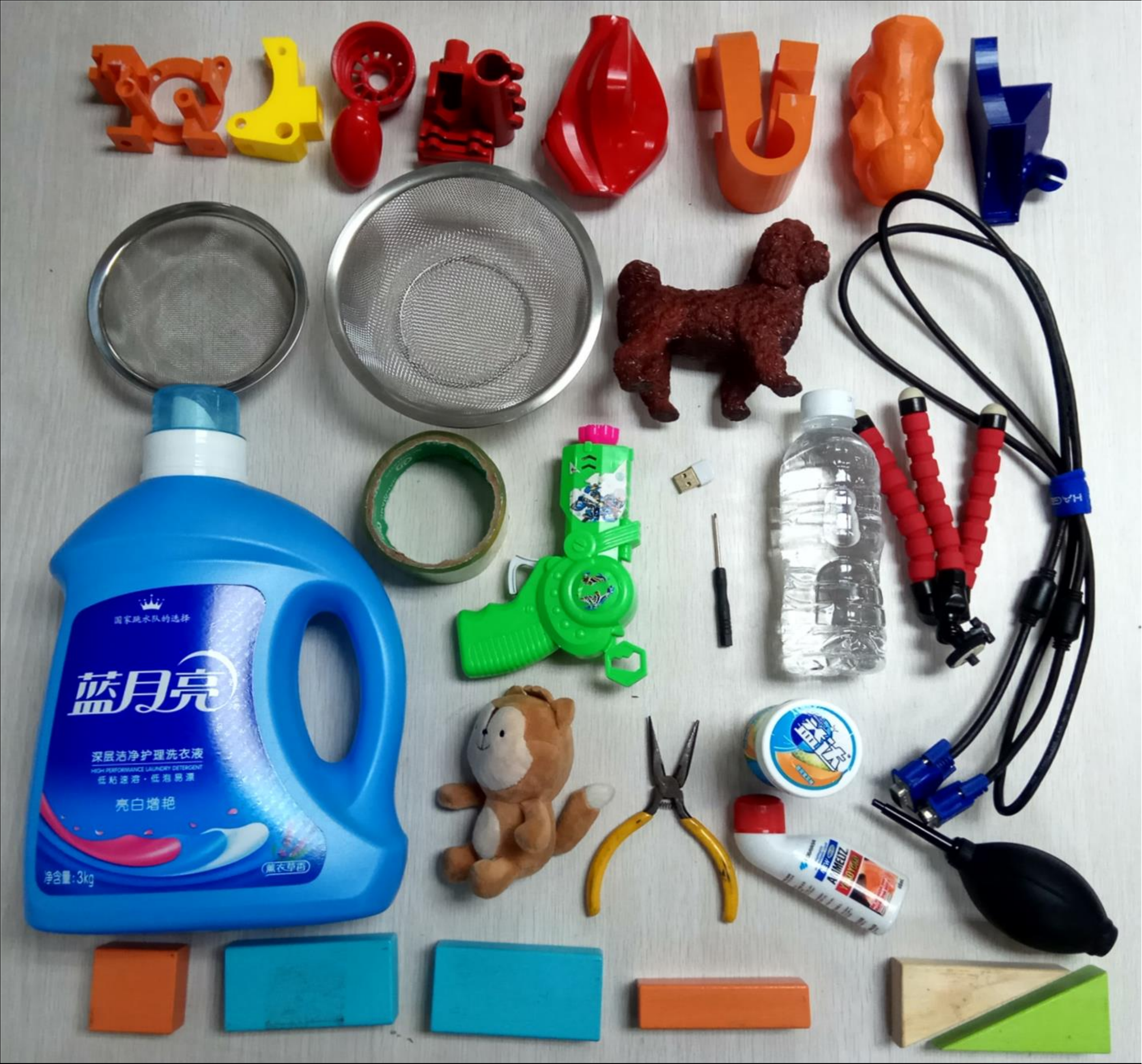}
    }
    \hspace{-11pt}
	\subfigure[Scenarios]{
        \label{fig:scenarios}
		\includegraphics[scale=0.231]{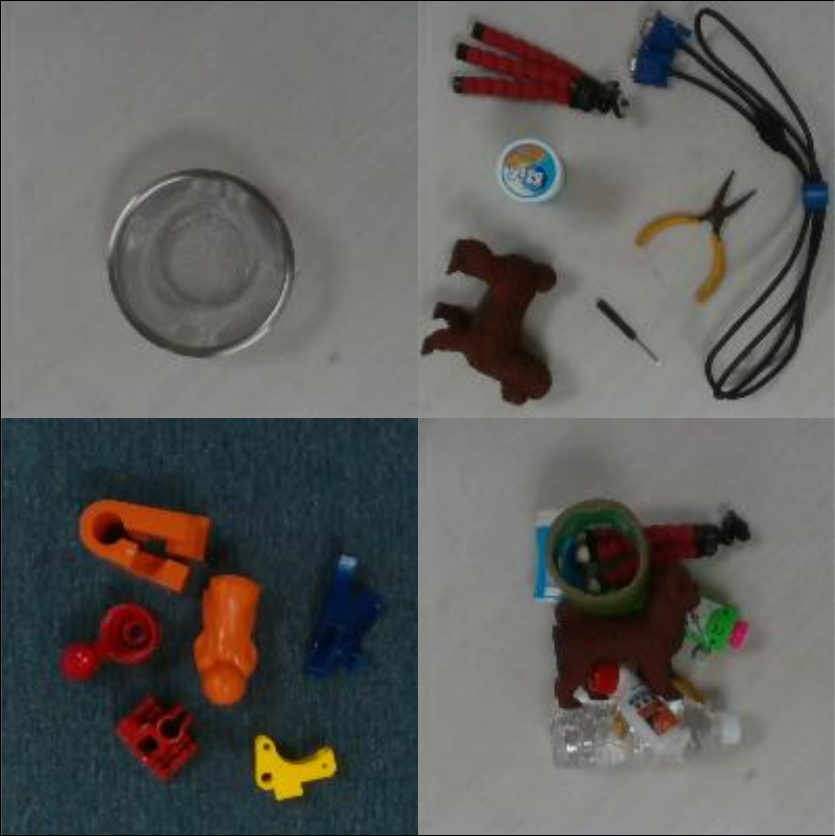}
    }
    \hspace{-11pt}
    \subfigure[Textures]{
        \label{fig:background textures}
		\includegraphics[scale=0.231]{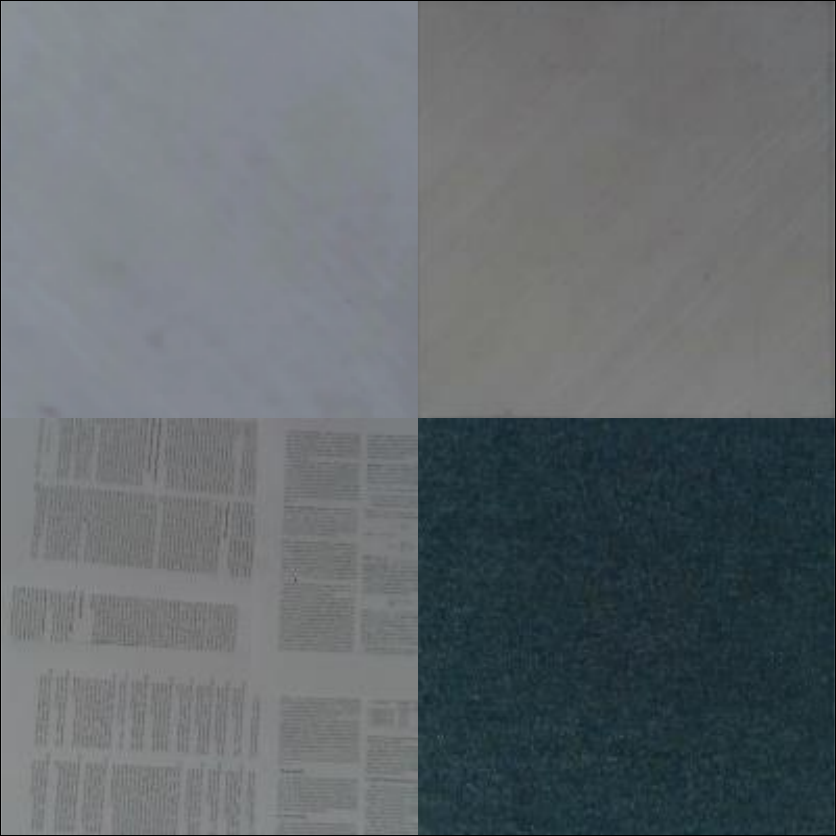}
    }
    \vspace{-10pt}
	\caption{{\bf{Grasp objects, scenarios and background textures.}} (a) Training objects. (b) Test objects. (c) The figures correspond to singular-object, multiple-object, cluttered and texture-change scenarios respectively in clockwise direction. (d) The first texture is used as background of simulated workspace, others are used in real world. There are background texture, texture1, texture2 and texture3 respectively in clockwise direction.}
\vspace{-20pt}
\end{figure}

\textbf{Training Dataset.} The training dataset is collected using totally 35 objects from 3DNet~\cite{wohlkinger20123dnet} in V-REP simulator~\cite{rohmer2013v}. The set of objects can be seen in Fig.~\ref{fig:train objects}. We totally collect 6370 samples in which there are 3264 positive grasp samples and 3106 negative samples respectively.

\textbf{Training Details.} We train our network with stochastic gradient descent method and set learning rate 0.001 with exponential decay and batch size 8. We augmented our data with vertical flip and slight rotation. For each sample, we rotate data to the left at most 5 degrees, so as to the right. Therefore, we augment our data to create a set of 140,140 samples.

\textbf{Test Objects.} We divide our test objects into three sets. One is household set containing 16 items of different sizes, shapes and physical properties. Another one is adversarial set consisting of 8 3D-printed items with adversarial geometry which were used by~\cite{mahler2017dex} and~\cite{morrison2018closing}. The last one is building blocks which have regular shapes, we use these objects to facilitate our qualitative analysis. All of the test objects are novel for our model. The objects are shown in Fig.~\ref{fig:test objects}.

\textbf{Scenario Configuration.} We evaluate the model performance with four different scenarios. The first is singular-object scenario, where there is only one object placed in the workspace. Once the gripper successfully grasps the object, it randomly rotates the object and place it to another position. The second one is multiple-object scenario, in which we can test whether the model is able to predict robust grasp configuration under multiple objects scene. The third one is clutter scene with 10 household objects. And the fourth one is similar to the first scene except the background texture. The scenarios are presented in Fig.~\ref{fig:scenarios}. In this experiment we can demonstrate the robustness of our proposed approach.

\vspace{-5pt}
\subsection{Qualitative Results}
\begin{figure}[t]	
	\begin{center}
		\includegraphics[scale=0.15]{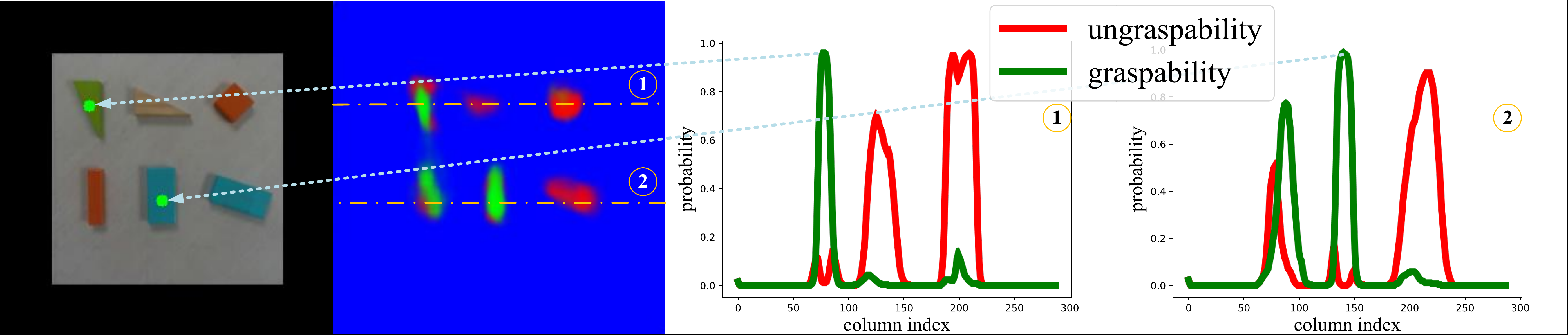}
	\end{center}
    \vspace{-10pt}
	\caption{{\bf{Horizontally antipodal grasp analysis.}} Top: Input the RGB image, the model outputs the pixelwise horizontal grasp affordance map. Buttom: Probability curves of two rows of affordance map.}
\label{fig:horizontally antipodal grasp analysis}
\vspace{-20pt}
\end{figure}
\textbf{Analysis of Horizontally Antipodal Grasp Pattern.} We predict affordance map given an RGB image in which there are 6 objects placed. The result is shown in Fig.~\ref{fig:horizontally antipodal grasp analysis}. We select two rows from affordance map and plot graspability and ungraspability curves of them. Row one shows that the positions of objects are able to be correctly located and labelled as non-background region. As for row two, we can see that the blue rectangle block placed vertically achieves higher probability of graspability, while the slant one in the lower right is prone to be ungraspable. The result demonstrates that the model can correctly find antipodal grasp pattern in horizontal direction.

\begin{figure}[t]	
	\begin{center}
		\includegraphics[scale=0.45]{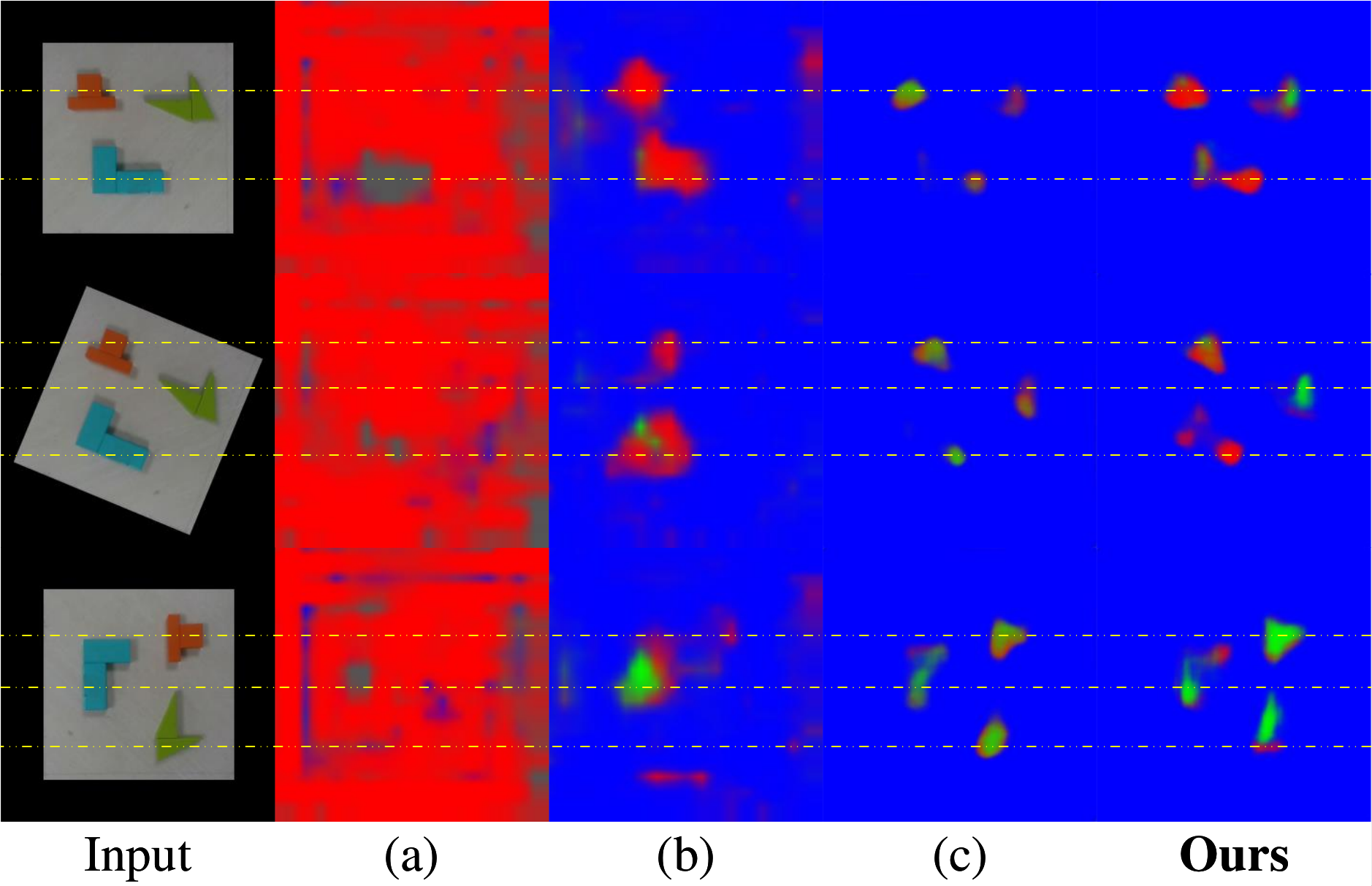}
	\end{center}
    \vspace{-10pt}
	\caption{{\bf{Grasp location analysis.}} Input the RGB image, the models predict affordance map respectively. Red represents ungraspable, green denotes graspable and blue is background. (a) is the result  of the model implemented according to~\cite{zeng2017robotic}, (b) is also the result of~\cite{zeng2017robotic} except the training loss. Column (c) is the affordance maps generated by the model trained with data collected by random grasp rule. The last column is our result. It shows that our approach can evaluate more accurate grasp location.}
\label{fig:Grasp location analysis}
\vspace{-20pt}
\end{figure}

\textbf{Analysis of Grasp Location.} Fig.~\ref{fig:Grasp location analysis} compares between our approach and other methods visually. We train a variant of model using the technique proposed by~\cite{zeng2017robotic} with our collected data. In~\cite{zeng2017robotic}, the model was trained with 0 loss propagation for the background regions, and the shape of affordance map is 8 times smaller than that of input image, hence the output is upsampled 8 times using bilinear interpolation to map input image. We can see that affordance map predicted by~\cite{zeng2017robotic} is ambiguous (Fig.~\ref{fig:Grasp location analysis}(a)), and most of background regions are misclassified. Therefore, we modify the training loss to retain background loss, while the shape of affordance map remains unchanged. The predicted affordance map is shown in Fig.~\ref{fig:Grasp location analysis}(b), it is coarse due to the interpolation and has some errors of grasp location. We also train the model using our proposed method except the data collected by random grasp rule instead of antipodal heuristic. The result is presented in Fig.~\ref{fig:Grasp location analysis}(c). We can see that the model predicts some positions where the objects are not perpendicular to the horizontal direction as graspable region. By contrast, our proposed network architecture is able to predict more accurate grasp location.

\vspace{-4pt}
\subsection{Quantitative Results}
We execute extensive grasp experiments in physical world. Some of the visual results are shown in Fig.~\ref{fig:affordance maps}. More sufficient results can be seen in the \emph{supplementary video}.

\begin{table}[h]
\scriptsize
    \vspace{-5pt}
    \caption{Singular-object grasp success rate}
    \vspace{-5pt}
    \label{singular-object success rate}
    \centering
    \begin{tabular}{ c | c | c | c | c | c | c}
    \hline
    Set&MAG&~\cite{mahler2017dex}&Ours(No CGS)&Ours(T1)&Ours(T2)&Ours(T3)\\
    \hline
    \hline
        H&77$\pm$0&-&85$\pm$6&\textbf{93$\pm$6}&92$\pm$3&90$\pm$4\\
    \hline
        A&80$\pm$6&93&89$\pm$8&91$\pm$3&92$\pm$7&\textbf{94$\pm$1}\\
    \hline
    \end{tabular}
    \vspace{1pt}
    \footnotesize{Note: H means household set and A is Adversarial Set. T1, T2 and T3 represent texture1, texture2 and texture3 respectively.}\\
    \vspace{-10pt}
\end{table}

\textbf{Singular Object Grasp.} In singular object experiment, we execute 20 grasps on each object and then average grasp success. For the household set, we achieve \textbf{93\%} success rate with 95\% confidence intervals. For the adversarial set, the accuracy obtained is \textbf{91\%} with 95\% confidence intervals. We also implement two variants of our approach, one is the multi-affordance grasping proposed by~\cite{zeng2017robotic} (MAG), another is the same as our method but without corrective grasp strategy (CGS) for data collection. The results are presented in Table~\ref{singular-object success rate}. MAG achieves 77\% and 80\% on household and adversarial items respectively, which shows that our pixel-level affordance interpreter network is able to predict more accurate grasp configuration. Moreover, although the model trained without corrective grasp strategy also achieves good performance, the visualization of the affordance maps presented in Fig.~\ref{fig:random grasp rule} and Fig.~\ref{fig:antipodal grasp rule} shows that model trained with data under antipodal grasp rule predicts more precise affordance than that under random grasp rule.

\begin{figure}[b]	
\vspace{-10pt}
    \centering
	\subfigure[Affordance maps]{
        \label{fig:affordance maps}
		\includegraphics[scale=0.175]{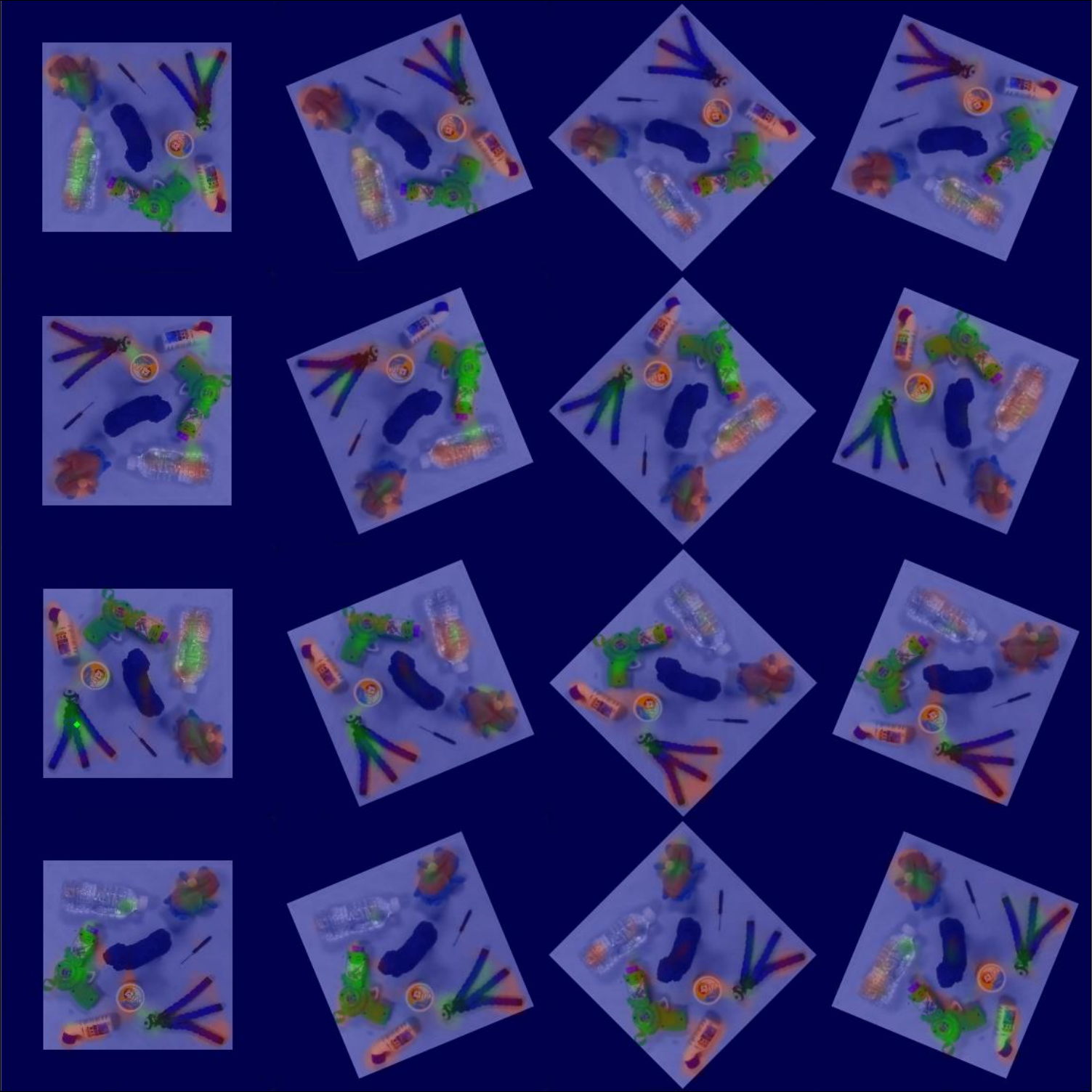}
    }
    \hspace{-10pt}
    \subfigure[Random grasp rule]{
        \label{fig:random grasp rule}
		\includegraphics[scale=0.35]{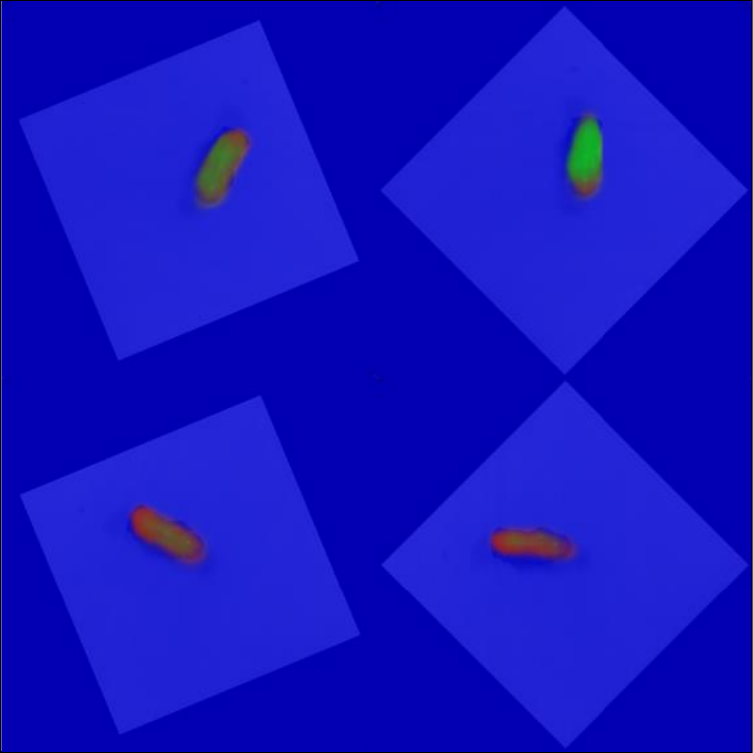}
    }
    \hspace{-10pt}
    \subfigure[Antipodal grasp rule]{
        \label{fig:antipodal grasp rule}
		\includegraphics[scale=0.35]{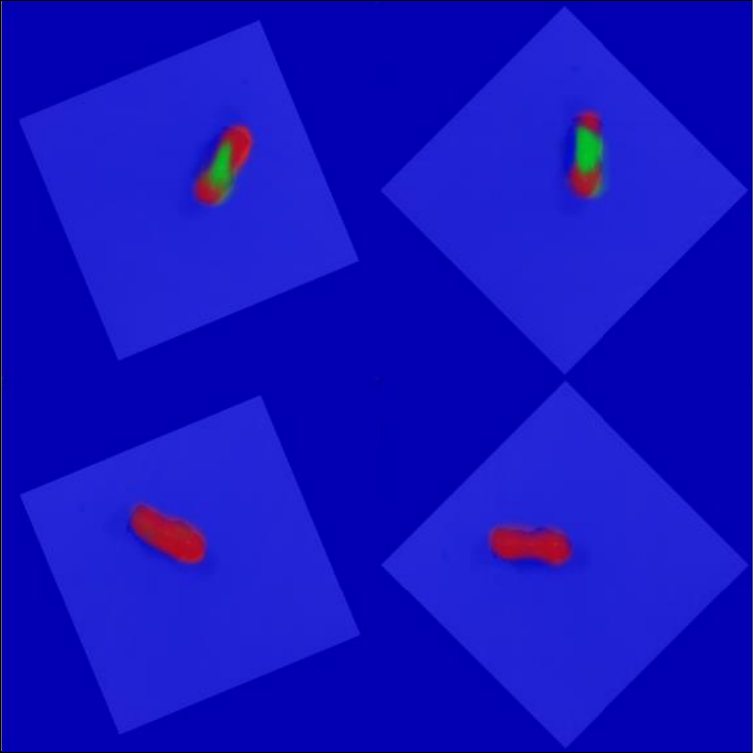}
    }
    \vspace{-10pt}
	\caption{{\bf{Visualization of affordance maps.}} The affordance maps of random grasp rule in (b) is ambiguous when the object is slant (i.e., there exists both red and green colors on the position of the object). Conversely, the results in (c) show that the model trained with data under antipodal rule is able to predict the slant object as ungraspability more accurately. Best viewed in color.}
\end{figure}

\textbf{Multiple Objects Grasp in Isolation.} In multiple objects experiments, we randomly select 6 objects placed on the workspace isolatedly and grasp them into the bin successively. This task repeats 10 times. As a result, we achieve \textbf{90\%} (60/67) success rate on household set and \textbf{94\%} (60/64) on adversarial set. Although our training data only contains singular object scenario, we can achieve significant performance on the multiple-object scene.

\textbf{Grasp in Clutter.} We attempt 10-time tries at removing 10 household items cluttered on the workspace. The 10 objects are first shaken in a box and emptied in a pile on the workspace. The above configuration is the same as~\cite{morrison2018closing}. Despite our data collection does not involve objects in clutter, we demonstrate that our model performs well not only on the objects in isolation but also on clutter scenario. We achieve \textbf{87\%} (93/107) success rate.

\textbf{Robustness to Different Backgrounds.} To demonstrate the robustness of our approach, we perform 20 grasp tries for each object under different background textures. We do grasp experiments on 3 different textures which are illustrated in Fig.~\ref{fig:background textures}. The success rate can be seen in Table~\ref{singular-object success rate}. The results show that our approach is robust to different background textures and that the model is able to effectively pay attention to learning antipodal grasp pattern as well as ignores the disturbing information of the RGB image.

\section{Conclusion and Future Work}

In this work, we propose a new grasping data collection method guided by antipodal grasping rule in virtual environment. The collected data can effectively contain horizontal antipodal grasp pattern. We apply only about 6,300 simulated samples to an end-to-end fully convolutional network to predict pixelwise grasp affordance map. Extensive experiments show that our proposed approach can achieve equivalent performance compared with current state-of-the-art methods. In this work we place emphasis on finding grasp pattern in synthesized RGB image, while there are also limitations to RGB image. In future work we will make the most of both RGB and depth information.

\bibliographystyle{IEEEtran}
\bibliography{egbib}
\end{document}